\documentclass[11pt]{article}

\usepackage[margin=1in]{geometry}
\usepackage{setspace}
\usepackage{float}
\usepackage{placeins}

\usepackage{parskip}              

\usepackage{newtxtext,newtxmath}  

\usepackage{graphicx}
\usepackage[font=small,labelfont=bf]{caption}
\usepackage{subcaption}
\usepackage{booktabs}
\usepackage{siunitx}
\usepackage{fancyhdr}
\usepackage{tikz}
\usepackage{enumitem}
\usepackage{float}
\usetikzlibrary{positioning, arrows.meta}

\usepackage{amsmath}
\usepackage{mathtools}
\usepackage{tabularx}
\usepackage{ragged2e}
\usepackage{graphicx}
\usepackage{subcaption}
\usepackage{caption}   

\usepackage[numbers]{natbib}
\usepackage[hidelinks]{hyperref}

\newcommand{\figcredit}[1]{\par\smallskip\textit{Graphic created by the student researcher using #1, \the\year.}\par}
\newcommand{\tablecredit}[1]{\par\smallskip\textit{Table created by the student researcher using #1, \the\year.}\par}
\newcommand{\Ipres}{\mathrm{I}_{\sim s}}
\newcommand{\Tpres}{T_{\Ipres}}

\begin{document}

\begin{titlepage}
\centering
{\Large \textbf{ABM-UDE: Developing Surrogates for Epidemic Agent-Based Models via Scientific Machine Learning}\par}
\vspace{1cm}

\noindent
\begin{minipage}[t]{0.47\textwidth}
\centering
{\large Sharv Murgai\par}
{\small Monta Vista High School\par}
{\small \texttt{murgai.sharv@gmail.com}\par}
\end{minipage}
\hfill
\begin{minipage}[t]{0.47\textwidth}
\centering
{\large Utkarsh Utkarsh\par}
{\small MIT CSAIL\par}
{\small \texttt{utkarsh5@mit.edu}\par}
\end{minipage}

\vspace{0.8cm}

\noindent
\begin{minipage}[t]{0.47\textwidth}
\centering
{\large Kyle C. Nguyen\par}
{\small Sandia National Laboratories\par}
{\small \texttt{kcnguye@sandia.gov}\par}
\end{minipage}
\hfill
\begin{minipage}[t]{0.47\textwidth}
\centering
{\large Alan Edelman\par}
{\small MIT CSAIL\par}
{\small \texttt{edelman@mit.edu}\par}
\end{minipage}

\vspace{0.8cm}

\noindent
\begin{minipage}[t]{0.47\textwidth}
\centering
{\large Erin C. S. Acquesta\par}
{\small Sandia National Laboratories\par}
{\small \texttt{eacques@sandia.gov}\par}
\end{minipage}
\hfill
\begin{minipage}[t]{0.47\textwidth}
\centering
{\large Christopher Vincent Rackauckas\par}
{\small MIT CSAIL\par}
{\small \texttt{crackauc@mit.edu}\par}
\end{minipage}

\end{titlepage}

\begin{abstract}
Agent-based epidemic models (ABMs) encode behavioral and policy heterogeneity but are too slow for nightly hospital planning. We develop county-ready surrogates that learn directly from exascale ABM trajectories using Universal Differential Equations (UDEs): mechanistic SEIR-family ODEs with a neural-parameterized contact rate $\kappa_\phi(u,t)$ (no additive residual). Our contributions are threefold: we adapt multiple shooting and an observer-based prediction--error method (PEM) to stabilize identification of neural-augmented epidemiological dynamics across intervention-driven regime shifts; we enforce positivity and mass conservation and show the learned contact-rate parameterization yields a well-posed vector field; and we quantify accuracy, calibration, and compute against ABM ensembles and UDE baselines. On a single representative ExaEpi scenario, PEM--UDE reduces mean MSE by 77\% relative to single-shooting UDE (3.00 vs.\ 13.14) and by 20\% relative to MS--UDE (3.75). Reliability improves in parallel: empirical coverage of ABM $10$--$90$\% and $25$--$75$\% bands rises from 0.68/0.43 (UDE) and 0.79/0.55 (MS--UDE) to 0.86/0.61 with PEM--UDE and 0.94/0.69 with MS+PEM--UDE, indicating calibrated uncertainty rather than overconfident fits. Inference is in seconds on commodity CPUs (20--35\,s per $\sim$90-day forecast), enabling nightly ``what-if'' sweeps on a laptop. Relative to a $\sim$100\,CPU-hour ABM reference run, this yields $\sim$10$^{4}$\,$\times$ (about 4 orders of magnitude) lower wall-clock per scenario. This closes the realism--cadence gap and supports threshold-aware decision-making (e.g., maintaining ICU occupancy $<75\%$), aligning analytic turnaround with daily data refreshes. The framework preserves mechanistic interpretability, delivers calibrated forecasts at operational speed, and runs on standard institutional hardware, making reliable, risk-aware scenario planning feasible for local health systems. Beyond epidemics, the ABM$\to$UDE recipe provides a portable path to distill agent-based simulators into fast, trustworthy surrogates for other scientific domains.
\end{abstract}

\newpage
\pagenumbering{arabic} 


\section{Introduction and Background}

\subsection{Motivation and Context}

During the COVID-19 pandemic, hospitals and county health systems faced a dual challenge:
rapidly evolving outbreaks demanded real-time planning, yet the analytic tools capable of
capturing human behavior and spatial heterogeneity, agent-based models (ABMs) , were
computationally prohibitive to run at the frequency decision-makers required \citep{SandiaSAND2024}, \citep{KitsonLoimos2024}. Agent-based modeling (ABM) represents systems as collections of autonomous, interacting agents whose local rules and interactions generate emergent, system-level dynamics \citep{bonabeau2002abm,macal2010tutorial,railsback2019abm}, a “bottom-up” approach popularized in social science and beyond \citep{epstein1996growing}, and used in infectious-disease modeling to capture realistic contact networks and interventions \citep{eubank2004nature}.
Hospitals made critical choices on the basis of partial or lagged information: staffing rosters,
elective-surgery postponements, inter-facility transfers, and surge-unit activation were often
reactive rather than pre-emptive. When intensive-care units (ICUs) approached capacity, mortality
spiked dramatically. According to a national CDC analysis, weeks in which ICU occupancy reached
roughly 75\% were followed by an estimated 12{,}000 excess deaths, while occupancy exceeding
100\% corresponded to approximately 80{,}000 excess deaths two weeks later
\citep{french2022impact}. In the Department of Veterans Affairs hospital system, patient mortality
was nearly twice as high during periods of heavy ICU load (75-100\% of peak) as during periods of
low demand \citep{BravataJNO2021}. These findings illustrate the concrete societal stakes of
forecast timeliness: \emph{every day of lead time} can translate into lives saved.

\subsection{Limitations of Current Modeling Paradigms}

Classical compartmental models based on systems of ordinary or partial differential equations
(ODEs/PDEs) are computationally efficient and interpretable but rely on strong assumptions of
homogeneous mixing and average rates. They cannot represent household clustering, workplace
interactions, or school-specific behaviors that drive local surges. In contrast, ABMs explicitly
represent individuals, their contacts, and policy-driven changes in behavior, making them uniquely
suited for forecasting hospital demand at county or facility resolution. The realism comes at a
computational price. For example, the \emph{CityCOVID} model \citep{SandiaSAND2024}, representing the 2.7~million
residents of the Chicago metropolitan area, requires approximately 100~CPU-hours for a
70-day simulation on conventional hardware, scaling to about 129~CPU-hours for a 90-day
scenario \citep{SandiaSAND2024}. The \emph{Loimos} framework simulates a 200-day outbreak on a
California digital twin (about 35~million agents) in 42~seconds by exploiting 4{,}096~CPU cores on
NERSC's Perlmutter supercomputer, achieving 4.6~billion traversed edges per second
\citep{KitsonLoimos2024}. Such throughput is unattainable outside HPC centers. Calibration of
these models is equally demanding: a 2024 Argonne campaign executed more than 32{,}000 runs of
CityCOVID using the ``Theta'' system, consuming roughly 420{,}000~core-hours across 6{,}400~cores
\citep{robertson2024b_rfcalib}. These numbers make clear that while ABMs are the gold standard for
epidemic realism, they remain inaccessible for most local public-health agencies.

\subsection{Forecast Cadence and Operational Needs}

For real-world decision support, not all forecasts are equally valuable.
Evaluations of COVID-19 mortality forecasts in the United States revealed that predictive skill
declines sharply with horizon: 20-week forecasts exhibit three to five times the error of one-week
forecasts \citep{cramer2022evaluation}. 
This implies that \textbf{frequent, near-term, localized forecasting} is vastly more actionable than
occasional long-range projections.
In practice, county health officers, hospital networks, and emergency planners need to evaluate
dozens of ``what-if'' scenarios each night, for example, ``What happens if mobility increases by
10\% next week?'', ``When should we open a field hospital?'', or ``How will ICU load change if
elective procedures resume on Monday?''.
Existing ABMs cannot provide such responsiveness: even a single 70-day CityCOVID run at
100~CPU-hours becomes infeasible when multiplied across hundreds of policy combinations.
Consequently, many local systems either rely on coarse-grained state-level models or operate
without quantitative scenario analysis.

\subsection{Research Gap}

The computational gap between ABM fidelity and operational cadence motivates the search for
hybrid modeling approaches.
An ideal framework would combine the interpretability and mechanistic grounding of
compartmental models with the contextual realism of ABMs, while achieving run times compatible
with daily hospital decision cycles.
To date, no such framework has been adopted at the county or health-system level.
Even when surrogate models or metamodels are proposed, they often target national-scale
averages or depend on HPC for inference.
Bridging this gap requires a paradigm that \emph{learns from ABMs} yet \emph{runs like ODEs}.

\subsection{Universal Differential Equations as a Bridge}
A \textbf{Universal Differential Equation (UDE)} couples known mechanistic equations with a learned
component while preserving biological and physical constraints. Throughout the paper, we use “UDE” to mean a \emph{neural-parameterized ODE} in which the contact rate $\kappa(t)$ is learned as $\kappa_\phi(u,t)$ \emph{inside} the mechanistic right-hand side; 
we do not add a free residual term $g_\phi$ to $\dot u$. Formally, let $\dot{u}=f(u,t;\theta)$ denote the mechanistic core (e.g., SEIR-family dynamics).
We replace the uncertain contact-rate profile $\kappa(t)$ by a bounded neural map
$\kappa_\phi(u,t)$, yielding
\begin{equation}
  \dot{u} \;=\; f\!\big(u,t;\theta,\,\kappa_\phi(u,t)\big),
  \qquad 
  \kappa_\phi(u,t) \;=\; \kappa_{\max}\,\sigma\!\big(N_\phi(\tilde u(t))\big),
\end{equation}
where $N_\phi$ is a small neural network, $\sigma$ is a squashing function (e.g., logistic), and
$\tilde u$ denotes normalized inputs. Because only the contact-rate coefficient within the infection
term is learned, the flow structure enforces positivity and mass conservation, preserving
interpretability of $\theta$ while allowing $\kappa_\phi$ to capture behavior-driven regime shifts,
reporting effects, or spatial spillovers evident in ABM or real-world data.
Practically, this keeps inference low-dimensional and fast (solving only a small ODE system), and
modern stiff-aware solvers with multiple-shooting or prediction-error optimization enable stable
training on noisy ABM trajectories or county-level series.

\subsection{Objectives, Hypotheses, and Expected Impact} \label{sec:objectives_hypotheses_impact}

This study aims to develop and evaluate a county-ready ABM$\rightarrow$UDE surrogate pipeline
that delivers ABM-level predictive accuracy at a fraction of the computational cost.
Specific objectives are:

\begin{itemize}
  \item[\textbf{O1.}] \textbf{Pipeline development:} Construct a training pipeline that ingests ABM-generated
  or observational data, fits a stiff-aware UDE using hybrid mechanistic-neural training, and
  provides under-a-minute inference for 70–90-day scenarios on commodity hardware.
  \item[\textbf{O2.}] \textbf{Validation and calibration:} Compare UDE forecasts to ABM trajectories and
  historical hospitalization/ICU data to quantify predictive accuracy and uncertainty.
  \item[\textbf{O3.}] \textbf{Operational demonstration:} Implement nightly scenario sweeps for representative
  counties or hospital systems, linked to capacity thresholds (e.g., ICU~$<\!75\%$), and measure
  lead-time gains for intervention decisions.
  \item[\textbf{O4.}] \textbf{Resource quantification:} Record runtime and core-hours
  to quantify computational and environmental savings.
\end{itemize}

\vspace{0.5em}
\noindent
The corresponding hypotheses are:

\begin{itemize}
  \item[\textbf{H1.}] Runtime per scenario will be at least $10^3$ times lower than ABM
  baselines (e.g., CityCOVID's $\sim$100\,CPU-h/70\,d and Loimos's dependence on 4{,}096~cores)
  \citep{SandiaSAND2024,KitsonLoimos2024}.
  \item[\textbf{H2.}] UDE surrogates will reproduce ABM aggregate behavior within operational
  tolerances for short-horizon (1–3~week) forecasts, consistent with where predictive skill is
  highest~\citep{cramer2022evaluation}.
  \item[\textbf{H3.}] The UDE surrogate will be computationally efficient enough to run thousands of
  “what-if” scenarios nightly on commodity hardware, enabling planners to identify policies that
  keep ICU occupancy below 75\%, a level associated with steep mortality gradients~\citep{french2022impact,BravataJNO2021}.
\end{itemize}

\vspace{0.5em}
\noindent
\textbf{Expected Impact:}
By compressing the computational intensity of ABMs into the compact, interpretable form of a UDE,
this work enables high-fidelity epidemic forecasting for \emph{any} health jurisdiction, not only those with supercomputing resources.
Running thousands of scenarios overnight allows officials to assess trade-offs among interventions, staffing, and transfer strategies
to keep hospitals below critical occupancy.
More broadly, the framework provides a blueprint for converting complex agent-based or simulation-heavy systems into
fast, physics-informed surrogates for real-time decision-making.

\textbf{Adaptation to Agent-Based Models:}
Multiple shooting, prediction-error methods, and universal differential equations are established in
system identification and scientific machine learning.
Our contribution lies in their \emph{adaptation} to the agent-based modeling (ABM) setting:
\begin{itemize}
    \item UDEs~\citep{rackauckas2020universal} serve as surrogates for exascale ABMs, replacing parameters such as
    the contact rate $\kappa(t)$ with neural terms capturing emergent, heterogeneous effects.
    \item Multiple shooting~\citep{tamimi2015multiple} is reformulated for neural-augmented epidemiological models,
    stabilizing training across abrupt interventions.
    \item PEM~\citep{chesebro2025scientific} is extended by interpolating averaged ABM trajectories and jointly
    learning observer gains $K$ with UDE parameters, anchoring the surrogate to stochastic ABM outputs while preserving differentiability.
\end{itemize}

Together, these adaptations bridge local ABM dynamics and tractable compartmental ODEs,
enabling scalable, mechanistically faithful surrogate modeling that supports the study's objectives and hypotheses.

\section{Related Work}

\textbf{Hybrid Scientific ML with Universal Differential Equations (UDEs).} augment mechanistic systems with learned residual terms that capture unmodeled effects while preserving structure and constraints \cite{rackauckas2020universal}. In epidemics, neural parameterizations of contact/transmission terms can absorb behavioral heterogeneity without sacrificing epidemiological interpretability. However, existing surrogates often emphasize national or coarse-grained settings and/or still rely on HPC for training or inference, which limits county-level adoption.
By training a stiff-aware UDE surrogate directly on ABM trajectories, our approach targets county-scale fidelity with commodity-hardware inference (\textbf{O1}) and enables rigorous cross-checks against ABM and hospital data (\textbf{O2}, \textbf{H2}). \textbf{Backpropagation} through long horizons with single shooting can accumulate numerical/estimation error and become brittle around intervention shocks. Multiple shooting (MS) improves stability by partitioning trajectories and softly enforcing continuity, a staple in system identification \cite{bock1984multiple,peifer2005parameter,tamimi2015multiple}. Observer/prediction-error methods repeatedly re-anchor states to measurements and can enhance calibration \cite{kalman1961bucy,ljung1998system}. Despite their maturity, these tools remain underused in epidemic UDE surrogates.
Our MS-UDE and PEM-UDE variants adapt these techniques to neural-augmented SEIR-family systems, addressing instability and calibration limits to support \textbf{O2}/\textbf{H2} while retaining the lightweight inference required by \textbf{O3}/\textbf{H1}. In summary, prior approaches are either (i) fast but too coarse (homogeneous-mixing ODEs) or (ii) realistic but too slow/expensive for nightly decision support (ABMs and HPC-dependent calibrations). Hybrid surrogates exist but often lack county-scale focus, training stability across dataset shifts, interpretability. By (a) learning a UDE surrogate from ABMs, (b) stabilizing training with MS and observer-based PEM, (c) grounding evaluation in short-horizon accuracy/coverage, (d) tying forecasts to ICU thresholds, and (e) quantifying compute, our work addresses these gaps and maps to \textbf{O1--O4} and \textbf{H1--H3} (see §\ref{sec:objectives_hypotheses_impact}).

\section{Methods}
\subsection{Overview of Approach}
\label{sec:overview}

\begin{figure}[t]
    \centering
    \includegraphics[width=\linewidth]{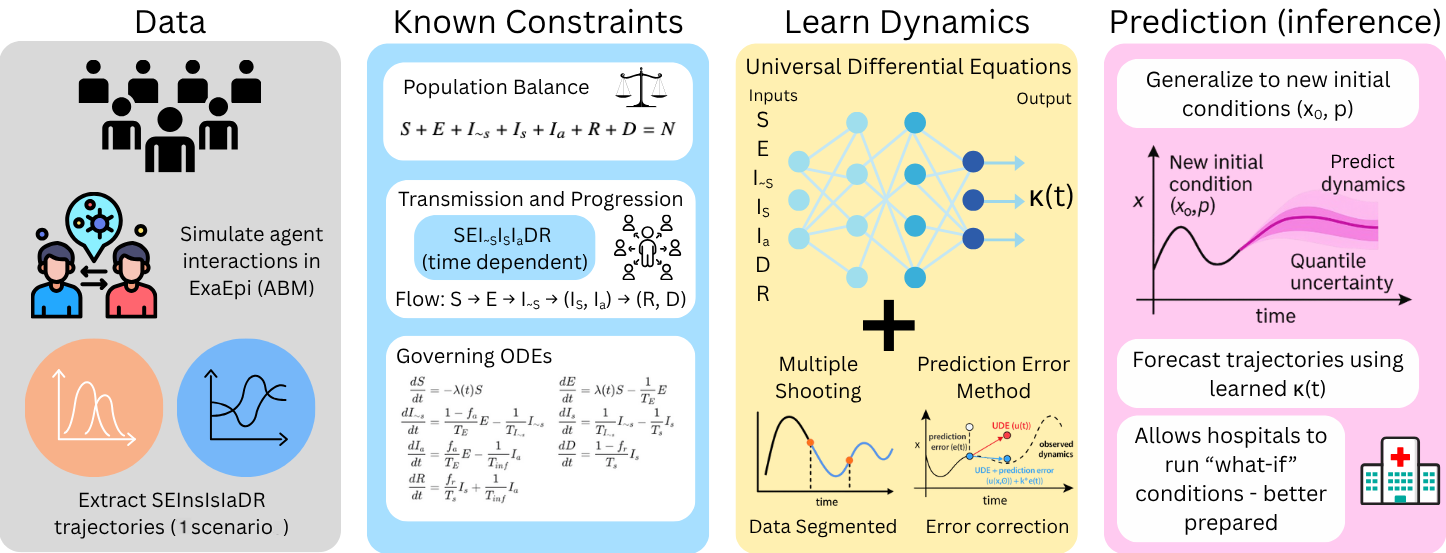}
    \caption{\textbf{Pipeline overview (one-scenario evaluation).}
    \emph{Data:} ExaEpi (ABM) generates a single representative SEInsIsIaDR outbreak trajectory (Dataset 1).
    \emph{Known constraints:} population balance, compartmental flow, and mechanistic ODE structure constrain learning.
    \emph{Learn dynamics:} a neural-parameterized UDE is trained with stabilization strategies: Vanilla UDE, MS-UDE,
    PEM-UDE, and MS+PEM-UDE.
    \emph{Ensemble robustness:} each method is trained as a 100-seed ensemble to quantify sensitivity to initialization
    and optimization stochasticity.
    \emph{Inference:} the trained UDE supports fast “what-if” forward simulation by varying $x_0$ and contact schedules
    $\kappa(t)$ on commodity hardware.}
    \label{fig:pipeline}
\end{figure}

Figure~\ref{fig:pipeline} outlines the framework integrating agent-based data with mechanistic and neural
components into a constrained Universal Differential Equation (UDE) surrogate. The workflow spans ABM data
generation, constraint enforcement, stabilized dynamics learning, and fast predictive inference.

In this report we center evaluation on \textbf{one representative ExaEpi scenario}
(\textbf{Dataset 1}), an AMReX-based agent simulation that produces daily SEInsIsIaDR compartment counts under
a fixed intervention schedule. While ExaEpi can generate many intervention regimes, multi-scenario evaluation
is outside the scope of this report.

Mechanistic realism is enforced through SEInsIsIaDR structure and invariants,
including the population balance
\begin{equation} \label{eq:population}
S + E + I_{\sim s} + I_s + I_a + R + D = N,
\end{equation}
nonnegativity of all compartments, monotonic $D(t)$, and bounded $\kappa(t)\in[0,1]$. The flow
$S \rightarrow E \rightarrow I_{\sim s} \rightarrow (I_s,I_a) \rightarrow (R,D)$ ensures interpretable transitions.
The governing ODE provides a mechanistic backbone with time-dependent transmission $\lambda(t)$ modulated by a
learned contact factor $\kappa(t)$ (see §\ref{sec:baseline-ode}, Eq.~\eqref{eq:lambda-phi}).

UDE learning is used to make sure missing dynamics are captured by learning $\kappa(t)$ as a bounded neural map
$\kappa_\phi(u,t)$ inside the mechanistic right-hand side (no additive residual), preserving positivity,
mass conservation, and interpretability.

To ensure stabilized training, we train four variants: (i) Vanilla UDE (single shooting),
(ii) MS-UDE (multiple shooting with continuity penalties across windows),
(iii) PEM-UDE (observer-based prediction error correction), and
(iv) MS+PEM-UDE, which applies the PEM-style correction within each shooting window while
still enforcing inter-window continuity. These strategies are designed to reduce long-horizon drift and
improve robustness around intervention-driven regime shifts.

For robustness, we use an ensemble (training uncertainty). To quantify sensitivity to random initialization and
optimization stochasticity, we train a 100-seed ensemble for each method and report ensemble mean
trajectories and variability bands (e.g., $\pm 1\sigma/\pm 3\sigma$) on Dataset 1.

Once trained, the UDE surrogate enables rapid forward simulation on commodity hardware.
In general, the framework can be used to explore counterfactual “what-if” interventions by varying initial
conditions $x_0$ and contact schedules $\kappa(t)$; in this report, results are reported for Dataset 1
scenario and its ensemble-based calibration analysis.

\subsection{Dataset / Simulation Source}
\label{sec:data}

All experiments in this report use one representative ExaEpi scenario (Dataset~1), generated with
\textsc{ExaEpi}, an exascale agent-based epidemiological simulator built on AMReX~\citep{githubGitHubAMReXAgentExaEpi}.
The scenario spans a fixed horizon of $K=89$ days with daily sampling and includes an intervention
change-point at $t_{\mathrm{ld}}=49$ (aligned with the step change in $\kappa(t)$ defined in
§\ref{sec:baseline-ode}). We use a single well-mixed region and report results only for this scenario.

For Dataset~1, \textsc{ExaEpi} simulates stochastic interactions among agents and outputs population counts for
the seven compartments $(S,E,I_{\sim s},I_s,I_a,R,D)$ at each day $t_k$.
To reduce Monte Carlo noise and obtain smooth training targets, we generate $R$ independent ABM realizations
for the \emph{same} scenario and form the ensemble mean trajectory $\bar{y}(t_k)$ for each compartment.
For calibration evaluation (§\ref{sec:eval_metrics}), we also compute empirical quantiles
$y^{(q_1)}(t_k)$ and $y^{(q_2)}(t_k)$ (e.g., 10th--90th percentiles) across these $R$ realizations.
The $D$ compartment is recorded as a \emph{cumulative} total.

To ensure numerical consistency with the ODE (§\ref{sec:baseline-ode}), we verify that
$S+E+I_{\sim s}+I_s+I_a+R+D \approx N$ at each $t_k$. If small discrepancies arise from stochastic rounding,
we apply a simplex projection to the living compartments to enforce mass conservation (see §\ref{sec:pem}).
All series are normalized by $N_0:=N-D(0)$ so that values lie in $[0,1]$.

We store the dataset in a wide time-series format with one row per day $t_k$ containing the 7D compartment
vector. Optional operational observables (\texttt{Hospitalized}, \texttt{ICU}, \texttt{Ventilated}) may be
recorded for threshold-oriented discussion, but they are not ODE state variables. Unless stated otherwise,
the observation operator is the identity on the seven compartments, $H=I_7$, so the losses in
§\ref{sec:ude}--§\ref{sec:ms} supervise all compartments (variance-weighted; Eq.~\ref{eq:ude-loss},
Eq.~\ref{eq:ms-data}), consistent with the evaluation metrics in §\ref{sec:eval_metrics}.

\subsection{Baseline Compartmental Model} \label{sec:baseline-ode}
This section describes the compartmental model structured as an ODE system. The model captures the population-level dynamics that emerge from microlevel agent behaviors simulated in the ExaEpi agent-based model. \citep{nguyen2024model,keeling2008modeling,anderson1991infectious,hethcote2000mathematics}. We consider a compartmental ODE system describing the progression of individuals through various stages of a COVID-19 infection. Individuals begin in the susceptible compartment \(S\), where they can become infected and transition to the exposed compartment \(E\) at a time-dependent infection rate \(\lambda(t)\). Following a latent period \(T_E\), exposed individuals become infectious but remain without symptoms. These individuals are classified as presymptomatic (\(I_{\sim s}\)) or asymptomatic (\(I_a\)) \citep{keeling2008modeling,anderson1991infectious,hethcote2000mathematics}. Presymptomatic individuals remain in this state for a period \(T_{I_{\sim s}} = T_{inc} - T_E\), where \(T_{inc}\) denotes the incubation period or the time from infection to symptom onset. Once this period ends, presymptomatic individuals move to the symptomatic compartment \(I_s\) \citep{he2020temporal,lauer2020incubation}. Throughout this model, it is assumed that the total population remains constant, accounting for both living and deceased individuals. (see \eqref{eq:population})

Let \( u(t) = [S, E, I_{\sim s}, I_s, I_a, R, D]^\top \) and define the feasible set 
\begin{equation}
\mathcal{X}_N = \{\, u \ge 0 \mid S + E + I_{\sim s} + I_s + I_a + R + D = N \,\}.
\end{equation}
Under the assumptions \(T_E, T_{I_{\sim s}}, T_s, T_{inf} > 0\) and \(0 \le f_a, f_r \le 1\), the system of ODEs below is \emph{forward invariant} on \(\mathcal{X}_N\); that is, if \(u(0) \in \mathcal{X}_N\), then \(u(t) \in \mathcal{X}_N\) for all \(t \ge 0\). 
Each compartment’s inflow/outflow is nonnegative and 
\(\frac{d}{dt}(S+E+I_{\sim s}+I_s+I_a+R+D)=0\), ensuring conservation of population. \citep{hethcote2000mathematics,jacquez1972compartmental}. 
The model tracks \textbf{seven compartments}: \textbf{S} (Susceptible individuals who have not yet been infected); \textbf{E} (Exposed individuals in the latent period, infected but not yet infectious); \textbf{$I_{\sim s}$} (Presymptomatic infectious individuals who will eventually develop symptoms); \textbf{$I_s$} (Symptomatic infectious individuals who may recover or die); \textbf{$I_a$} (Asymptomatic infectious individuals who never develop symptoms but can transmit); \textbf{D} (Deceased individuals due to COVID-19); and \textbf{R} (Recovered individuals with immunity after infection).

The governing ODEs are: \citep{keeling2008modeling,anderson1991infectious,hethcote2000mathematics}
\begin{align}
\frac{dS}{dt} &= -\lambda(t) S, \label{A.1} \\[6pt]
\frac{dE}{dt} &= \lambda(t) S - \frac{1}{T_E}E, \label{A.2} \\[6pt]
\frac{dI_{\sim s}}{dt} &= \frac{1 - f_a}{T_E}E - \frac{1}{T_{I_{\sim s}}} I_{\sim s}, \label{A.3} \\[6pt]
\frac{dI_s}{dt} &= \frac{1}{T_{I_{\sim s}}} I_{\sim s} - \frac{1}{T_s} I_s, \label{A.4} \\[6pt]
\frac{dI_a}{dt} &= \frac{f_a}{T_E} E - \frac{1}{T_{\text{inf}}} I_a, \label{A.5} \\[6pt]
\frac{dD}{dt} &= \frac{1-f_r}{T_s} I_s, \label{A.6} \\[6pt]
\frac{dR}{dt} &= \frac{f_r}{T_s} I_s + \frac{1}{T_{\text{inf}}} I_a. \label{A.7}
\end{align}

The durations $T_E$, $T_{I_{\sim s}}$, $T_s$, and $T_{\text{inf}}$ are characteristic
periods (days) governing transitions between states:
latent, presymptomatic, symptomatic, and asymptomatic infectious periods, respectively.
The parameters $f_a$ and $f_r$ denote the fractions of individuals who become
asymptomatic infectious and who recover from symptomatic infection (death fraction $1-f_r$).
Time is measured in days, and compartment values are normalized by 
$N_0$ for numerical stability.

To account for differing infectiousness across states, we define the force of infection 
\citep{keeling2008modeling,diekmann2000mathematical} as
\begin{equation}
\lambda(t)
  = p_t\,\kappa(t)\,
    \frac{\eta_p\, I_{\sim s}(t) + I_s(t) + \eta_a\, I_a(t)}
         {N - D(t)} ,
\label{eq:lambda-phi}
\end{equation}
\begin{equation}
\kappa(t) = \kappa_\phi\big(u(t),t\big).
\label{eq:kappa-nn}
\end{equation}

where
\begin{itemize}
  \item $p_t \in [0,1]$ is the per-contact transmission probability,
  \item $\kappa(t)$ is the time-dependent average daily contact rate,
  \item $\eta_p,\eta_a \in [0,1]$ weight presymptomatic and asymptomatic infectiousness relative to symptomatic (unit weight),
  \item $N$ is total population and $D(t)$ is cumulative deaths (so $N-D(t)$ is the living population).
\end{itemize}
Following \citet{nguyen2024model} we may adopt the convention $\eta_p=\eta_a$ when appropriate, but we keep symbols distinct to avoid ambiguity.

We treat the triplet $(T_E,\,T_{\text{inc}},\,T_{\text{inf}})$ as \emph{foundational} epidemiological inputs (latent, incubation, and total infectious periods, respectively), and define the remaining durations by
\begin{align}
  \Tpres &= T_{\text{inc}} - T_E, \label{eq:pre-sym}\\
  T_s    &= T_E + T_{\text{inf}} - T_{\text{inc}}. \label{eq:sym}
\end{align}
All rate/duration parameters satisfy $T_E,\,T_{I_{\sim s}},\,T_s,\,T_{\text{inf}} > 0$ with the natural constraint
$T_{\text{inc}} \ge T_E$, which implies $T_{I_{\sim s}}\ge 0$; moreover $T_s>0$ whenever
$T_{\text{inf}} > T_{\text{inc}} - T_E$. Fractions such as $f_a, f_r \in [0,1]$.
These constraints ensure well-posed, physically meaningful transitions in the compartmental flows
\citep{hethcote2000mathematics,he2020temporal,lauer2020incubation}. To model behavioral changes due to lockdown measures, the contact rate \(\kappa(t)\) is represented as a step change \citep{dehning2020inferring,flaxman2020estimating,nguyen2024model}:
\begin{equation} \label{eq:kappa-step}
    \kappa(t) =
    \begin{cases}
    \kappa_1, & \text{if } t < t_{\text{ld}},\\[2pt]
    \kappa_2, & \text{if } t \ge t_{\text{ld}}.
    \end{cases}
\end{equation}
Here, \(t_{\text{ld}}\) denotes the lockdown onset time (e.g., \(t_{\text{ld}}=49\) days in Dataset 1). This piecewise-constant form matches the intervention dataset used in the ABM and keeps the epidemiological interpretation explicit. 
For gradient-based training of neural-augmented models, some implementations temporarily replace the hard step by a smooth approximation (e.g., a short-width logistic transition) to improve differentiability of adjoint gradients. Our \emph{model specification} remains the step form in \eqref{eq:kappa-step}; any smoothing, if used, is an optimization aid only and does not change the reported model. We can prove well-posedness through the Picard–Lindelöf theorem \citep{coddington1956theory}. The right-hand side of Eqs.\ (1)–(7) is locally Lipschitz on \(\mathcal{X}_N\), therefore the system admits a unique continuously differentiable solution \citep{coddington1956theory} \(u(t)\) for all \(t \ge 0\). 

\subsection{Universal Differential Equation (UDE) Formulation}

Rather than adding a free-form residual to the right-hand side, we use a
\emph{neural-parameterized ODE} in which the contact-rate process is learned as a
bounded neural surrogate inside the mechanistic RHS (no additive residual).
Formally, given baseline dynamics
\begin{equation}
\dot{u} = f_\theta(u,t),
\end{equation}
we replace the uncertain contact-rate profile $\kappa(t)$ by a neural map
$\kappa_\phi(u,t)$ \citep{chen2018neural,rackauckas2020universal}, yielding
\begin{equation}
\dot{u} = f_\theta\!\big(u,t;\kappa_\phi(u,t)\big).
\label{eq:neural-param-ode}
\end{equation}
A practical parameterization enforces boundedness via a squashing function, e.g.
\begin{equation}
\kappa_\phi(u,t) \;=\; \kappa_{\max}\,\sigma\!\big(N_\phi(\tilde u(t))\big),
\label{eq:kappa-nn-form} 
\end{equation}
where $N_\phi$ is a small neural network, $\sigma$ is the logistic sigmoid, and
$\tilde u$ denotes the normalized input features. We omit $S$ from $\tilde u$ since $S(t)=N-D(t)-E(t)-I_{\sim s}(t)-I_s(t)-I_a(t)-R(t)$ by mass balance,
so it is redundant given the other components. This representation preserves the biological structure and conservation laws of the
SE--$I_{\sim s}$--$I_s$--$I_a$--R--D system, because only the contact-rate coefficient within the mechanistic infection term is learned, while flows and
compartment couplings remain mechanistic. Consequently, the parameters $\theta$
retain epidemiological meaning, and $\kappa_\phi$ is interpretable as an effective,
behavior-driven modulation inferred from ABM trajectories. In practice, $\phi$ (and, when applicable, selected components of $\theta$) are estimated by minimizing a data-misfit objective (e.g., Eq.~\ref{eq:ude-loss})
with gradient-based optimization \citep{chen2018neural,rackauckas2020universal}.

\subsection{Training Strategies}
To effectively train the hybrid mechanistic–neural model, we explore three complementary strategies that progressively enhance stability and accuracy: single-shooting as a baseline (\ref{sec:ude}), MS for long-horizon robustness (\ref{sec:ms}), and an observer-augmented PEM for adaptive correction (\ref{sec:pem}).

\subsubsection{Vanilla-UDE Baseline} \label{sec:ude}

As a baseline, we train the neural-parameterized ODE from Section 3.4
(Eq.~\ref{eq:neural-param-ode}) using standard \textbf{single-shooting optimization}. For the neural component $N_\phi$ (from Eq.~\ref{eq:kappa-nn-form}), we use a
feed-forward neural network (NN) with \textbf{three hidden layers of 10 units}
and \textit{Swish} activations \citep{ramachandran2017searching}. The network
takes the six normalized states $\tilde u$ as input, as defined in
Section 3.4. In the single-shooting approach, gradients are backpropagated through the
full simulation trajectory. While conceptually straightforward, this method can become unstable over long horizons, especially when intervention change-points induce abrupt changes within a trajectory
\citep{nguyen2024model}.

Observed data are modeled as
\begin{equation}
\mathbf{y}(t_k) = H\,\mathbf{u}(t_k) + \boldsymbol{\varepsilon}_k,
\label{eq:obs-model}
\end{equation}
where $H$ selects observed compartments and $\boldsymbol{\varepsilon}_k$ is
measurement noise \citep{ljung1998system}. We train the model parameters
$(\theta, \phi)$ by minimizing the following variance-weighted
mean-squared error (MSE) \citep{ljung1998system}:
\begin{align}
\mathcal{L}_{\mathrm{UDE}}(\theta,\phi)
&= \frac{1}{K}\sum_{k=1}^{K}
\big\|\,W\big(H\,\hat{\mathbf{u}}(t_k;\theta,\phi) - \mathbf{y}(t_k)\big)\big\|_2^2,
\label{eq:ude-loss}
\\
W &= \mathrm{diag}(w_1,\ldots,w_J), \quad w_j = 1/\hat{\sigma}_j. \nonumber
\end{align}
where $\hat{\mathbf{u}}(t_k;\theta,\phi)$ denotes simulated state at
time $t_k$, $J$ is number of observed series, and $\hat{\sigma}_j$
are scale estimates for normalization. This weighting prevents
large-magnitude states (e.g., $S$) from dominating  loss \citep{ljung1998system}.

\subsubsection{Multiple Shooting (MS)} \label{sec:ms}
To address instability in single shooting, we adapt a multiple-shooting (MS) \citep{tamimi2015multiple} scheme for UDE training. The time horizon is partitioned into $M$ contiguous windows $\mathcal{W}_i=\{s_i,\ldots,e_i\}$ of fixed length $m$ 
(observation indices). Each window solves a short initial-value problem (IVP); continuity across windows 
is enforced softly at the boundaries. \citep{bock1984multiple,diehl2005nominal,peifer2005parameter} This reduces the effective backpropagation horizon, improving stability 
under abrupt within-trajectory regime changes (e.g., the intervention at $t_{\mathrm{ld}}$). All neural and epidemiological parameters $(\phi,\theta)$ are \emph{shared} across windows; only the window 
initial states $\{z_i\}$ are window-specific. Although multiple shooting is a classical technique in numerical analysis, our contribution lies in formulating it for neural-augmented epidemiological models, where continuity constraints propagate through neural network parameters embedded in mechanistic dynamics.

Let $0=t_0<t_1<\cdots<t_T$ be observation times and partition the
index set $\{0,\dots,T\}$ into $M$ contiguous windows $\mathcal{W}_i=\{s_i,\dots,e_i\}$ of fixed length
$m$ (``group\_size'' in code). For each window we solve a short IVP and penalize discontinuities
between windows. We reuse the same learned contact rate (Eq.~\ref{eq:kappa-nn-form}), force of infection (from Eq.\ref{eq:lambda-phi}), and UDE formulation (Eq.~\ref{eq:neural-param-ode}). We now define the simulations and initial states for each window. For window $\mathcal{W}_i=\{s_i,\ldots,e_i\}$ we introduce an initial state $z_i\in\mathcal{X}_N$ at time $t_{s_i}$ 
($z_1$ is fixed to the known $u(0)$; subsequent $z_{i>1}$ may be optimized or anchored to data) \citep{jacquez1972compartmental,hethcote2000mathematics}. 
To ensure positivity and mass conservation by construction, we parameterize $z_i$ via
\begin{align}
\pi_i &= \mathrm{softmax}(w_i)\in\mathbb{R}^6,\qquad D_i = N\,\sigma(\delta_i), \nonumber\\
\big[S,E,I_{\sim s},I_s,I_a,R\big]_i &= (N-D_i)\,\pi_i, \qquad 
z_i=\big[S,E,I_{\sim s},I_s,I_a,R,D\big]_i. \label{eq:ms-simplex}
\end{align}
We then simulate $\hat{\mathbf{u}}_i(t;z_i,\theta,\phi)$ for $t\in[t_{s_i},t_{e_i}]$ by integrating \eqref{eq:neural-param-ode}. Similar to the vanilla UDE, observations are modeled as in Eq.~\eqref{eq:obs-model}: 
$\mathbf{y}(t_k)=H\,\mathbf{u}(t_k)+\boldsymbol{\varepsilon}_k$, 
where $H$ selects observed compartments. The per-window data misfit uses a variance-weighted MSE \citep{ljung1998system,peifer2005parameter}:
\begin{align}
L_{\mathrm{data}}(\theta,\phi,\{z_i\})
&= \sum_{i=1}^{M}\;\sum_{k\in \mathcal{W}_i}
\big\|W\big(H\,\hat{\mathbf{u}}_i(t_k;z_i,\theta,\phi)-\mathbf{y}(t_k)\big)\big\|_2^2,
\qquad W=\mathrm{diag(w_1,\ldots,w_J)},\; w_j=1/\hat{\sigma}_j. \label{eq:ms-data}
\end{align}
Global consistency is encouraged via a soft continuity penalty at window boundaries \citep{bock1984multiple,peifer2005parameter}:
\begin{align}
L_{\mathrm{cont}}(\theta,\phi,\{z_i\})
&= \lambda_{\mathrm{MS}}\sum_{i=1}^{M-1}\left\|
\hat{\mathbf{u}}_i(t_{e_i};z_i,\theta,\phi)-z_{i+1}\right\|_2^2, \label{eq:ms-cont}
\end{align}
where $\lambda_{\mathrm{MS}}>0$ is the continuity weight. Combining the data loss and continuity loss, the resultant multiple-shooting objective is
\begin{align}
\min_{\theta,\phi,\{z_i\}} \; L_{\mathrm{MS}}:=L_{\mathrm{data}}+L_{\mathrm{cont}}. \label{eq:ms-objective}
\end{align}
We use single-trajectory windows with shared $(\theta,\phi)$ across all windows and optimize $\{z_i\}$ 
subject to \eqref{eq:ms-simplex}. Training uses an Adaptive Moment Estimation (ADAM) \citep{kingma2014adam} warm start followed by Limited-memory Broyden–Fletcher–Goldfarb–Shanno (LBFGS) \citep{liu1989limited} refinement; 
we apply gradient clipping and stiff-aware solver tolerances (see §\ref{sec:impl}). Once again, to prove well-posedness of the multiple shooting differential equations, we see the right-hand side in \eqref{eq:neural-param-ode} is a composition of smooth, locally Lipschitz maps (smooth hidden layer activations and a final sigmoid activation, in \eqref{eq:kappa-nn-form}, linear $H$, and the mechanistic $f_\theta$), 
each windowed IVP is well-posed and admits a unique solution on $[t_{s_i},t_{e_i}]$. 
The reparameterization \eqref{eq:ms-simplex} guarantees $z_i\in\mathcal{X}_N$, and the infection term (from Eq.\ref{eq:lambda-phi}) preserves the bilinear structure and denominator $(N-D)$ used in §\ref{sec:baseline-ode}, 
so total population invariance is maintained within each window.

\subsubsection{Prediction Error Method (PEM)} \label{sec:pem}
Let \(\{(t_k,y_k)\}_{k=0}^{T}\) be observations with \(y_k\in\mathbb{R}^{J}\).
We model partial and noisy measurements by the same equation as in the UDE \eqref{eq:obs-model},
but here \(H\in\mathbb{R}^{J\times n}\) selects (and possibly aggregates) the observed compartments
and \(\varepsilon_k\) denotes measurement noise~\citep{ljung1998system}.
To align solver steps with data, we build a continuous, piecewise-linear interpolant
\(y:[t_0,t_T]\to\mathbb{R}^{J}\) through the samples \(\{(t_k,y_k)\}\),
as is standard in PEM implementations (cf. text above Eq.~\eqref{eq:obs-model}). In PEM, an additional observer corrects prediction errors.
Let \(\kappa_\phi\) be the learned contact rate from §3.4 and the mechanistic RHS \(f_\theta\) from §3.3.
We use a fixed-gain observer that corrects along the observed subspace
\citep{luenberger1964observing,kalman1961bucy,ljung1998system}:
\begin{equation}
\dot u \;=\; f_{\theta}\big(u,t;\kappa_{\phi}(u,t)\big)\;+\;K\,\big(y(t)-H\,u\big),
\qquad K\in\mathbb{R}^{n\times J}.
\label{eq:pem-predictor}
\end{equation}
A convenient parameterization that preserves this shape is
\[
K \;=\; D\,H^\top, \qquad D=\mathrm{diag}(k_S,\dots,k_D)\in\mathbb{R}^{n\times n},
\]
i.e., diagonal state gains \(D\) composed with \(H^\top\), which maps innovations in observation space to state corrections. To preserve interpretability and limit overfitting, we often constrain \(D\) to be diagonal (or block-diagonal).
The additive correction in \eqref{eq:pem-predictor} can push states slightly
outside the feasible set \(\mathcal{X}_N=\{u\ge0,\; \mathbf{1}^\top u=N\}\) defined in §\ref{sec:baseline-ode}.
We therefore integrate with a positivity- and mass-preserving projection \(\Pi_{\mathcal{X}_N}\) after each
solver step (or at fixed substeps):
\begin{equation}
u(t_{m+1}) \;\leftarrow\; \Pi_{\mathcal{X}_N}\!\big(u(t_{m+1}^{-})\big).
\label{eq:pem-proj}
\end{equation}
This projection enforces non-negativity and mass conservation (Euclidean simplex projection
in \(O(n\log n)\) time~\cite{Projection,condat2016fast}), keeping \(u\in\mathcal{X}_N\).

\paragraph{PEM objective.}
\textit{If targets are normalized by \(N_0\):}
\begin{align}
\mathcal{L}_{\mathrm{PEM}}(\theta,\phi,K)
&= \frac{1}{T+1}\sum_{k=0}^{T}
\Big\|\, W\!\left(H\,\frac{\hat u(t_k;\theta,\phi,K)}{N_0}-y_k\right)\!\Big\|_2^2
\;+\; \lambda_K \|K\|_F^2 \;+\; \lambda_{\mathrm{sp}} \|M\!\odot\!K\|_1,
\label{eq:pem-loss}
\end{align}
\textit{otherwise (targets in counts), drop the \(1/N_0\):}
\begin{align}
\mathcal{L}_{\mathrm{PEM}}(\theta,\phi,K)
&= \frac{1}{T+1}\sum_{k=0}^{T}
\Big\|\, W\!\left(H\,\hat u(t_k;\theta,\phi,K)-y_k\right)\!\Big\|_2^2
\;+\; \lambda_K \|K\|_F^2 \;+\; \lambda_{\mathrm{sp}} \|M\!\odot\!K\|_1.
\end{align}
Here \(\hat u(\cdot;\theta,\phi,K)\) is the solution of \eqref{eq:pem-predictor} with projection
\eqref{eq:pem-proj}, \(W=\mathrm{diag}(w_1,\dots,w_J)\) with \(w_j=1/\hat\sigma_j\),
\(M\in\{0,1\}^{n\times J}\) is an optional sparsity mask, and \(\lambda_K,\lambda_{\mathrm{sp}}\ge0\).

\noindent\textit{Well-posedness.}
Assume (i) \(f_\theta(\cdot,t;\kappa_\phi)\) is locally Lipschitz on \(\mathcal{X}_N\) (see §\ref{sec:baseline-ode}, §\ref{sec:ude}),
(ii) \(y(t)\) is piecewise-\(C^1\), and (iii) \(K\) is bounded. Then the RHS of \eqref{eq:pem-predictor}
is locally Lipschitz, so the observer admits a unique Carathéodory solution on \([t_0,t_T]\).
With \eqref{eq:pem-proj}, each step returns \(u\) to \(\mathcal{X}_N\), preserving nonnegativity and mass.

\subsubsection{MS+PEM: Multiple Shooting with Observer-Based Prediction Error Correction}
\label{sec:ms_pem}

Single-shooting UDE training can drift over long horizons, while MS shortens the backpropagation path by
segmenting trajectories (MS; §\ref{sec:ms}) and §\ref{sec:pem} re-anchors the dynamics to observations via an innovation term
(Sec.~3.5.3). We combine these ideas into a single stabilized estimator, \textbf{MS+PEM}, which applies an
observer-corrected UDE \emph{within each shooting window} while still enforcing global continuity across
windows. Intuitively, MS handles intervention-driven regime shifts by restarting integration periodically,
and PEM prevents within-window error accumulation by continuously correcting toward data.

\paragraph{Windowed observer dynamics.}
Let the observation times be $0=t_0<t_1<\cdots<t_T$ and partition indices into $M$ contiguous windows
$W_i=\{s_i,\ldots,e_i\}$ of fixed length $m$ as in Sec.~3.5.2. We use the same mechanistic RHS
$f_\theta(\cdot,t;\kappa_\phi)$ with neural contact-rate $\kappa_\phi(u,t)$ (Sec.~3.4). For each window $W_i$,
we introduce an initial state $z_i\in X_N$ at time $t_{s_i}$ and integrate an \emph{observer-augmented} UDE:
\begin{equation}
\dot u
=
f_\theta\!\big(u,t;\kappa_\phi(u,t)\big)
+
K\big(y(t)-Hu\big),
\qquad t\in[t_{s_i},t_{e_i}],
\label{eq:ms_pem_observer}
\end{equation}
where $H\in\mathbb{R}^{J\times n}$ selects observed compartments and $y(t)$ is a continuous interpolant of
$\{(t_k,y_k)\}_{k=0}^T$ (piecewise linear). We parameterize the gain as $K = D H^\top$ with diagonal (or
block-diagonal) $D=\mathrm{diag}(k_S,\ldots,k_D)$ to limit overfitting and preserve interpretability.

Because the additive correction can push states outside the feasible set
$X_N=\{u\ge 0:\mathbf{1}^\top u = N\}$, we enforce constraints by projection (as in Sec.~3.5.3). Concretely,
after each solver step (or at fixed substeps) we apply
\begin{equation}
u(t_{m+1}) \leftarrow \Pi_{X_N}\!\left(u(t^-_{m+1})\right),
\label{eq:ms_pem_projection}
\end{equation}
which restores nonnegativity and mass conservation.

\paragraph{MS+PEM objective: data fit + continuity + gain regularization.}
Let $\hat u_i(t;z_i,\theta,\phi,K)$ denote the solution of
\eqref{eq:ms_pem_observer}--\eqref{eq:ms_pem_projection} on window $W_i$ with initial state $z_i$.
We minimize a variance-weighted data misfit across all windows plus a soft continuity penalty between
adjacent windows:
\begin{align}
L_{\text{data}}
&=
\sum_{i=1}^M\ \sum_{k\in W_i}
\left\|
W\Big(H\hat u_i(t_k;z_i,\theta,\phi,K)-y_k\Big)
\right\|_2^2,
\label{eq:ms_pem_data}
\\
L_{\text{cont}}
&=
\lambda_{\text{MS}}
\sum_{i=1}^{M-1}
\left\|
\hat u_i(t_{e_i};z_i,\theta,\phi,K)-z_{i+1}
\right\|_2^2,
\label{eq:ms_pem_cont}
\\
L_{\text{gain}}
&=
\lambda_K\|K\|_F^2
+
\lambda_{\text{sp}}\|M\odot K\|_1,
\label{eq:ms_pem_gain}
\\
\min_{\theta,\phi,K,\{z_i\}}
\quad
L_{\text{MS+PEM}}
&:= L_{\text{data}} + L_{\text{cont}} + L_{\text{gain}}.
\label{eq:ms_pem_total}
\end{align}
Here $W=\mathrm{diag}(w_1,\ldots,w_J)$ with $w_j=1/\hat\sigma_j$ prevents large-magnitude series from
dominating, $\lambda_{\text{MS}}>0$ controls inter-window consistency, and $L_{\text{gain}}$ regularizes the
observer gain (optionally with a sparsity mask $M\in\{0,1\}^{n\times J}$).

\paragraph{Initial-state parameterization.}
As in Sec.~3.5.2, we ensure $z_i\in X_N$ by construction using a softmax-based parameterization:
\begin{equation}
\pi_i=\mathrm{softmax}(w_i)\in\mathbb{R}^6,\qquad
D_i=N\sigma(\delta_i),\qquad
(S,E,I_{\sim s},I_s,I_a,R)_i = (N-D_i)\pi_i,
\label{eq:ms_pem_zi}
\end{equation}
and set $z_i=[S,E,I_{\sim s},I_s,I_a,R,D]_i^\top$.

\paragraph{Why MS+PEM helps.}
Multiple shooting reduces sensitivity to intervention shocks by resetting integration on short windows,
while PEM keeps trajectories anchored to observed data throughout each window. Their combination
directly targets two dominant failure modes of single-shooting training: (i) long-horizon drift after small
early misfit, and (ii) instability around abrupt regime changes (e.g., lockdown-induced kinks). In practice,
MS+PEM often converges faster to lower error floors than MS alone because the observer reduces the burden
on the neural contact-rate $\kappa_\phi$ to ``explain away'' accumulated state errors.

\paragraph{MS+PEM-UDE ensemble study.}
We train MS+PEM-UDE with the same architecture and training budget as MS-UDE and PEM-UDE, and quantify
uncertainty by repeating training across 100 random seeds. We report ensemble mean trajectories and
$\pm1\sigma$ / $\pm3\sigma$ bands, and evaluate calibration using empirical coverage against ABM quantile
envelopes.

\subsection{Implementation Details } \label{sec:impl}

Models were implemented in the Julia programming language using the \texttt{DifferentialEquations.jl} ecosystem \citep{rackauckas2017differentialequations} for stiff-aware integration. We used a two-stage training procedure (ADAM followed by LBFGS) to minimize the mean-squared error across all observed compartments. All experiments were conducted on an Apple Macbook Air (M4, 10-core CPU, 16GB RAM).

\subsection{Evaluation Metrics}
\label{sec:eval_metrics}
Model performance is evaluated using complementary metrics that assess accuracy, stability, and efficiency, capturing both mechanistic fidelity and operational practicality. All metrics report mean~$\pm$~SD across independent runs with different random seeds. Note that the ABM ensemble (stochastic simulator replicates used to form quantile bands) and the surrogate training ensemble (100 random initializations used to form $\pm\sigma$ bands) are distinct uncertainty sources.

\begin{itemize}[leftmargin=9.5pt]
  \item \textbf{Trajectory accuracy:}  
  Mean-squared error (MSE)~\cite{friedman2001elements,goodfellow2016deep} from Eq.~\ref{eq:ude-loss}, averaged across all compartments, measures agreement between surrogate and ABM trajectories.
  
  \item \textbf{Uncertainty coverage:}  
  Fraction of predicted points lying within the ABM ensemble’s $q_1$–$q_2$ quantile band (e.g., 10–90\%)~\cite{gneiting2014probabilistic,cramer2022evaluation}:  
  \[
  \mathrm{Cov}_j = \frac{1}{K}\sum_{k=1}^{K}\mathbf{1}\!\left[
  y_j^{(q_1)}(t_k) \le \hat{u}_j(t_k) \le y_j^{(q_2)}(t_k)\right].
  \]
  Higher values indicate surrogate uncertainty consistent with ABM variability.

  \item \textbf{Continuity consistency:}  
  Normalized continuity penalty~\cite{bock1984multiple,tamimi2015multiple} (Eq.~\ref{eq:ms-cont}) quantifies inter-window smoothness in multiple-shooting training; lower values imply more stable segment transitions.

  \item \textbf{Computational efficiency:}  
  Wall-clock training time, forward solves, and solver iterations per epoch are recorded.  
  Relative speedups are reported against both single-shooting UDE and ExaEpi ABM baselines,
  highlighting runtime and core-hour reductions~\cite{ma2021comparison,henderson2020towards}.

  \item \textbf{Aggregate metrics:}  
  All metrics report mean~$\pm$~SD across independent runs with different random seeds and trajectory splits.
  Per-compartment trajectories are visualized against ABM quantile envelopes, with MSE, coverage, and continuity summarized in tables.
\end{itemize}

\section{Results}

\subsection{Trajectory Fit and Robustness (Ensemble Visualization)}
\label{sec:comparisons}

We evaluate surrogate fidelity and robustness on the representative scenario (Dataset~1) using a consolidated
ensemble visualization. Figure~\ref{fig:grid4x1_ms_pem} reports \textbf{100-seed training ensembles} for each
method, summarized by the ensemble mean and $\pm1\sigma/\pm3\sigma$ variability bands across random
initializations and optimizer stochasticity. This avoids over-interpreting any single-seed trajectory and
directly reflects stability under repeated training.

Across all compartments, stabilization tightens ensemble dispersion and improves agreement with ABM ground truth
(black \texttt{x} markers), particularly after the intervention change-point (vertical dotted line at day 49).
MS-UDE reduces long-horizon drift by shortening the effective backpropagation path via windowed training;
PEM-UDE further re-anchors the latent state using an observer correction; and MS+PEM-UDE combines both mechanisms.
Qualitatively, MS+PEM-UDE yields the most concentrated trajectories while remaining centered on the ground truth
through post-intervention dynamics.

\begin{figure}
    \centering
    \includegraphics[width=0.75\linewidth]{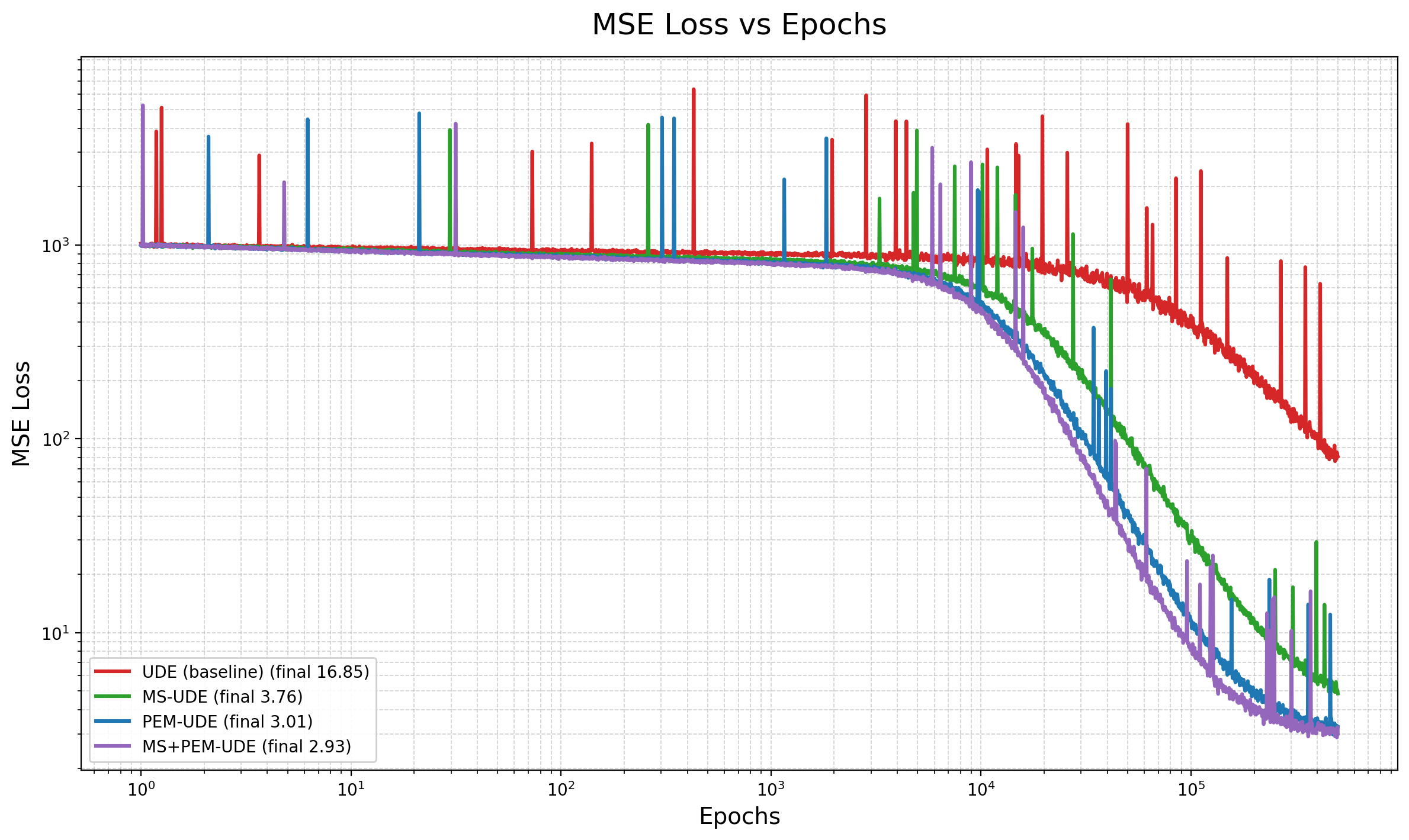}
    \caption{Epochs v MSE for all methods (100-seed avg)}
    \label{fig:loss-plot}
\end{figure}

\subsection{Quantitative Metrics}
\label{sec:metrics}

\paragraph{Overall trajectory error (MSE).}
Table~\ref{tab:mse_summary_dataset1} summarizes the final mean-squared error (MSE) on Dataset~1 for each method.
Stabilization yields a large reduction in error relative to the single-shooting baseline, and MS+PEM-UDE achieves
the lowest overall MSE among the evaluated approaches (Fig.~\ref{fig:loss-plot}).

\begin{table}[t]
  \centering
  \caption{Final mean-squared error (MSE) on Dataset 1 (average over 100 seeds). Lower is better.}
  \label{tab:mse_summary_dataset1}
  \begin{tabular}{lc}
    \toprule
    \textbf{Method} & \textbf{Final MSE (Dataset 1)} \\
    \midrule
    UDE (baseline) & 16.85 \\
    MS-UDE & 3.76 \\
    PEM-UDE & 3.01 \\
    MS+PEM-UDE & \textbf{2.93} \\
    \bottomrule
  \end{tabular}
\end{table}

\paragraph{Incremental improvement from MS+PEM over all other methods (per-compartment deltas).}
To quantify the added value of combining both stabilization mechanisms, Table~\ref{tab:mspem_all_deltas}
reports the relative change in per-compartment MSE of MS+PEM-UDE compared to PEM-UDE on Dataset~1
(negative values indicate MS+PEM-UDE is lower/better). These deltas correspond to the six compartments shown in
Fig.~\ref{fig:grid4x1_ms_pem}.

\paragraph{Inference runtime.}
Table~\ref{tab:inference_m4} reports per-trajectory inference latency for a $\sim$90-day horizon on an Apple
MacBook Air (M4). All UDE surrogates run in tens of seconds per forecast on a single CPU thread, enabling rapid
scenario evaluation relative to an ABM reference run.

\begin{table}[t]
  \centering
  \caption{Per-trajectory inference latency on Apple M4 (single thread, warmed session, daily \texttt{saveat}, moderate tolerances) compared with an ExaEpi ABM reference run. 
  Latency values are medians with $p90$ in parentheses (seconds for UDEs; hours for ABM). 
  The ABM reference corresponds to approximately 100 CPU-hours of execution.}
  \label{tab:inference_m4}
  \begin{tabular}{lc}
    \toprule
    \textbf{Method} & \textbf{Latency} \\
    \midrule
    UDE (baseline)         & \textbf{21}\,s\,(\,24\,s\,) \\
    MS-UDE (inference)     & \textbf{23}\,s\,(\,26\,s\,) \\
    PEM-UDE ($K\!\neq\!0$) & \textbf{31}\,s\,(\,35\,s\,) \\
    MS+PEM-UDE ($K\!\neq\!0$) & \textbf{32}\,s\,(\,35\,s\,) \\
    \addlinespace
    ExaEpi ABM (reference) & \textbf{100 h} = 360{,}000 s \\
    \bottomrule
  \end{tabular}
\end{table}

\newpage

\begin{center}
\captionsetup{type=figure} 
  \begin{subfigure}[t]{0.31\textwidth}
    \includegraphics[width=\linewidth]{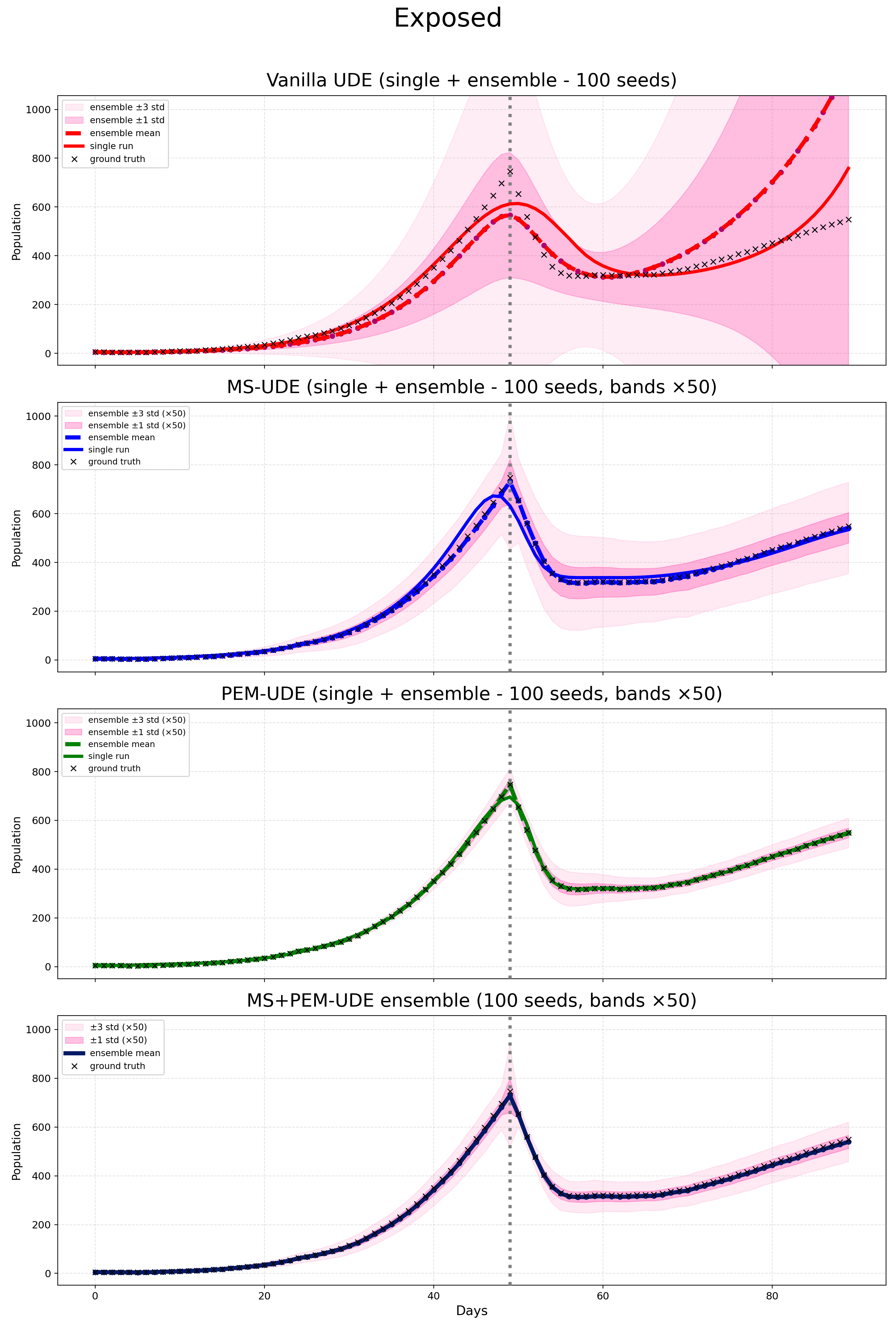}
    \caption{Exposed ($E$).}\label{fig:grid4x1_exposed}
  \end{subfigure}\hfill
  \begin{subfigure}[t]{0.31\textwidth}
    \includegraphics[width=\linewidth]{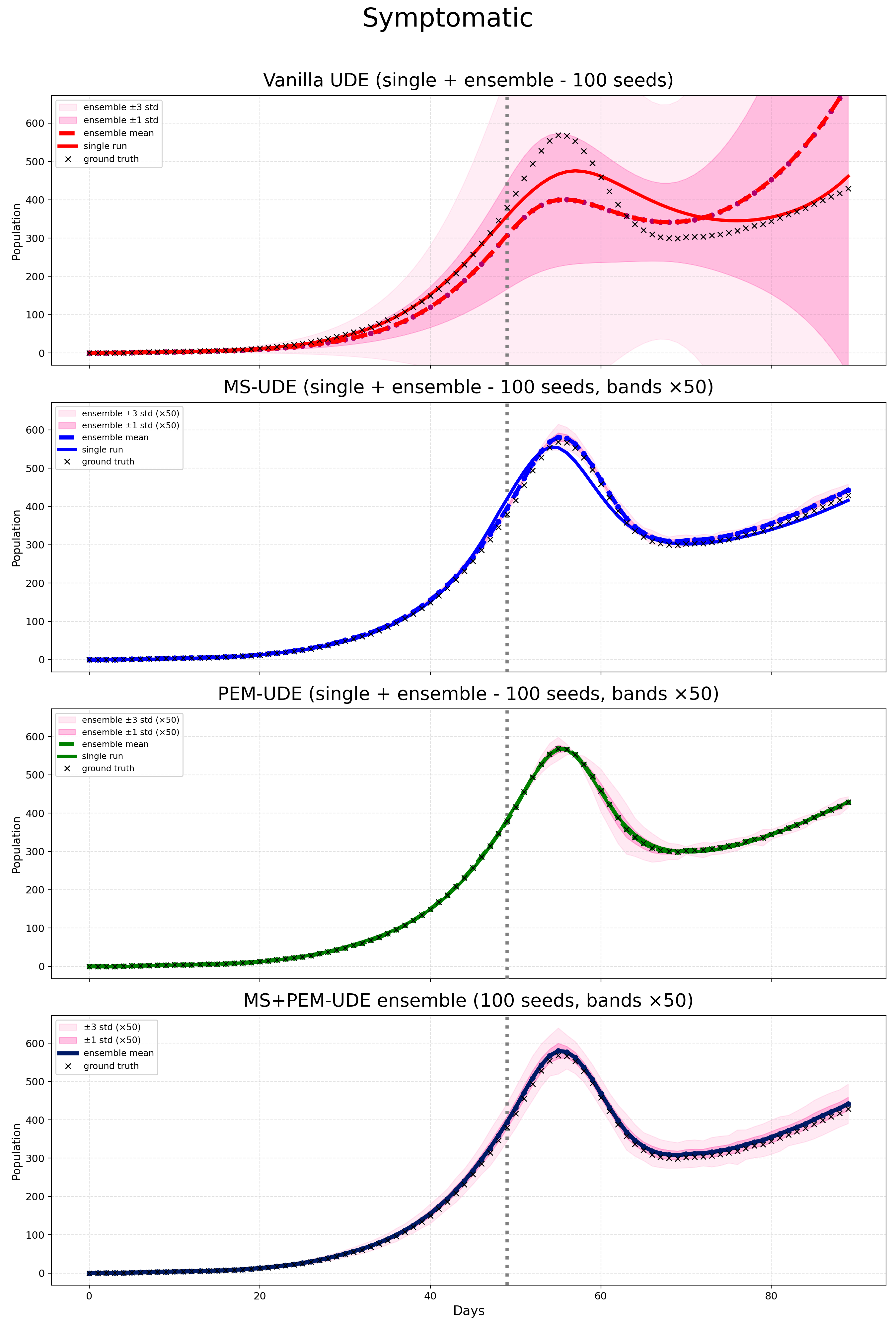}
    \caption{Symptomatic ($I_s$).}\label{fig:grid4x1_symptomatic}
  \end{subfigure}\hfill
  \begin{subfigure}[t]{0.31\textwidth}
    \includegraphics[width=\linewidth]{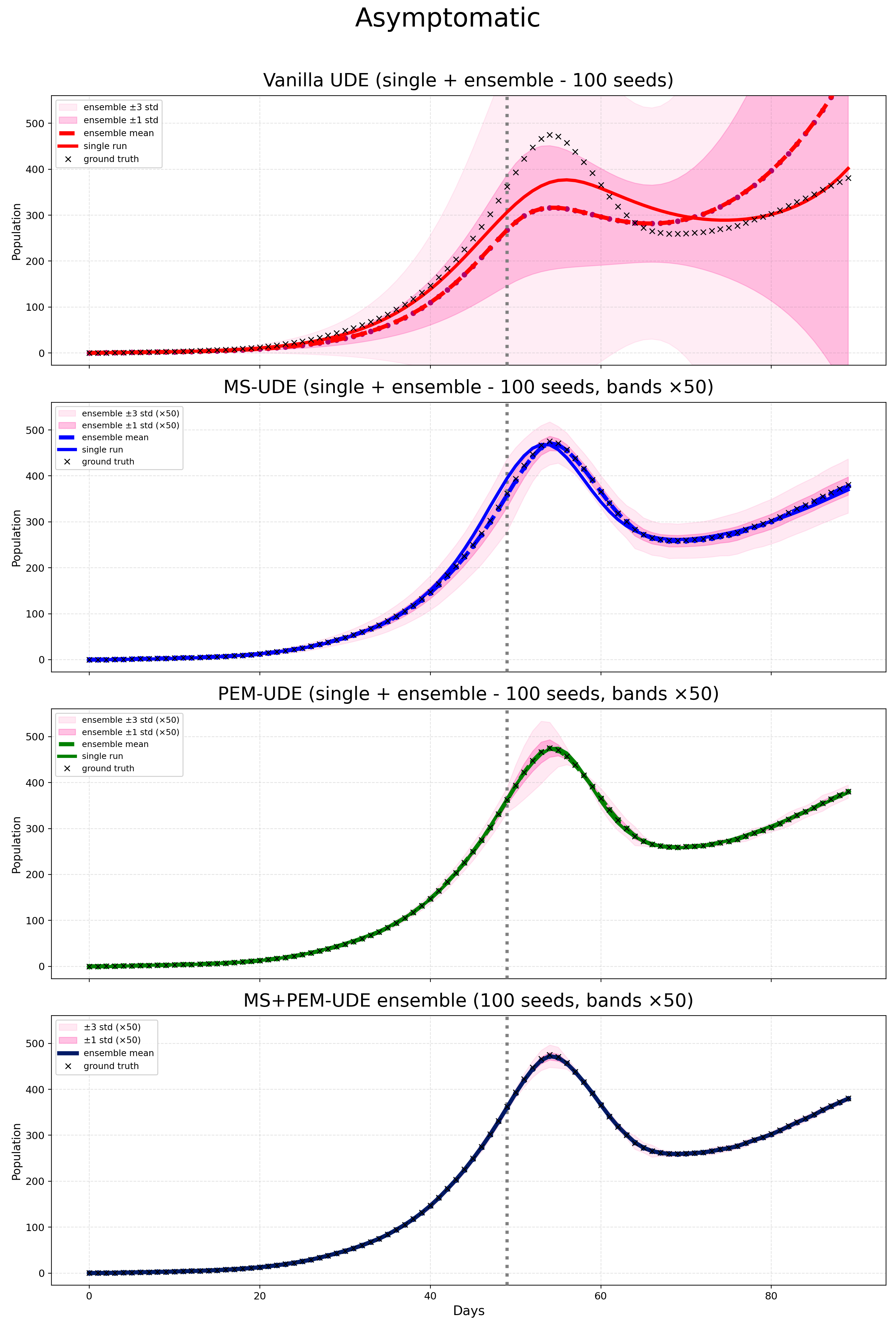}
    \caption{Asymptomatic ($I_a$).}\label{fig:grid4x1_asymptomatic}
  \end{subfigure}

  \vspace{0.3em}

  \begin{subfigure}[t]{0.31\textwidth}
    \includegraphics[width=\linewidth]{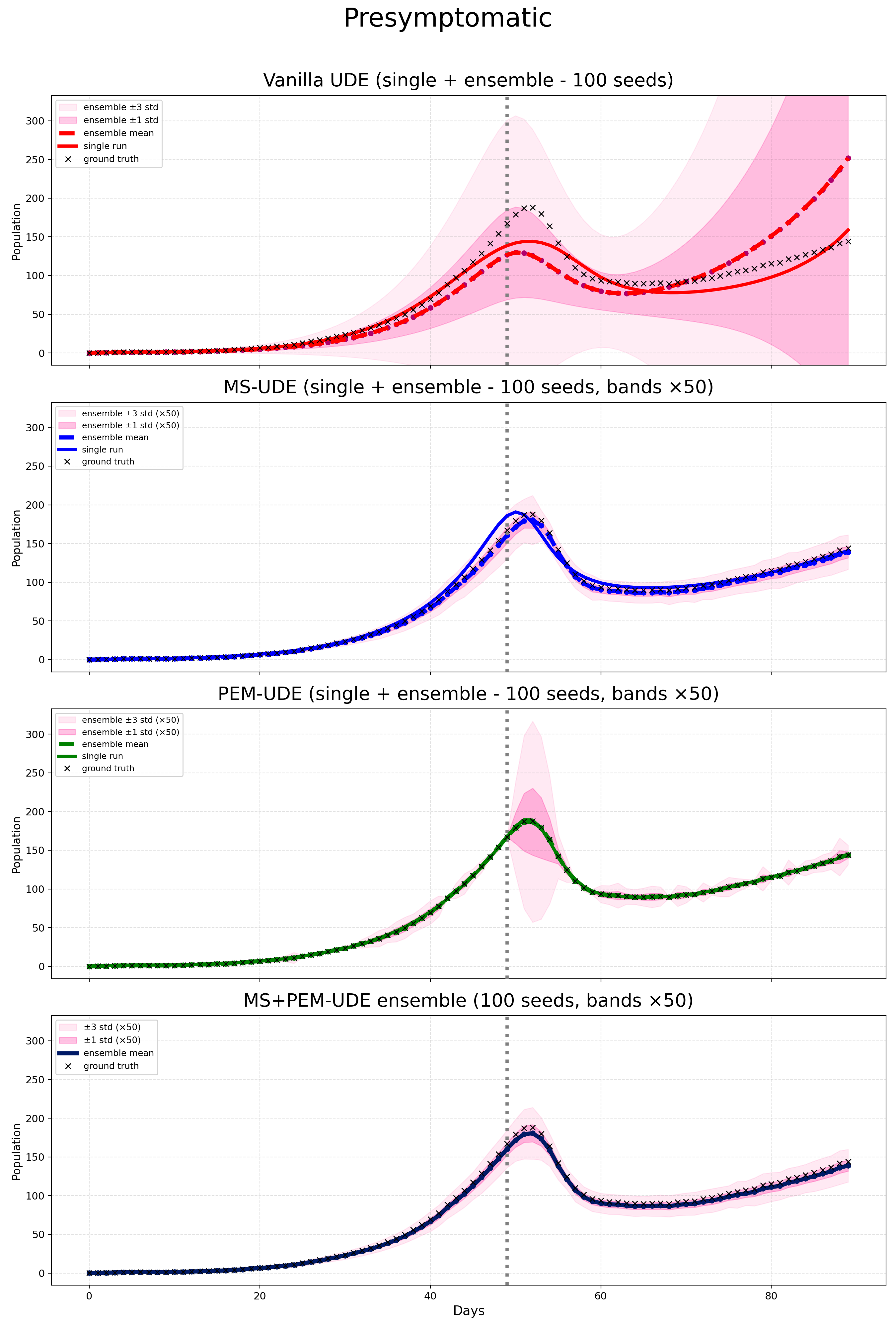}
    \caption{Presymptomatic ($I_{\sim s}$).}\label{fig:grid4x1_presymptomatic}
  \end{subfigure}\hfill
  \begin{subfigure}[t]{0.31\textwidth}
    \includegraphics[width=\linewidth]{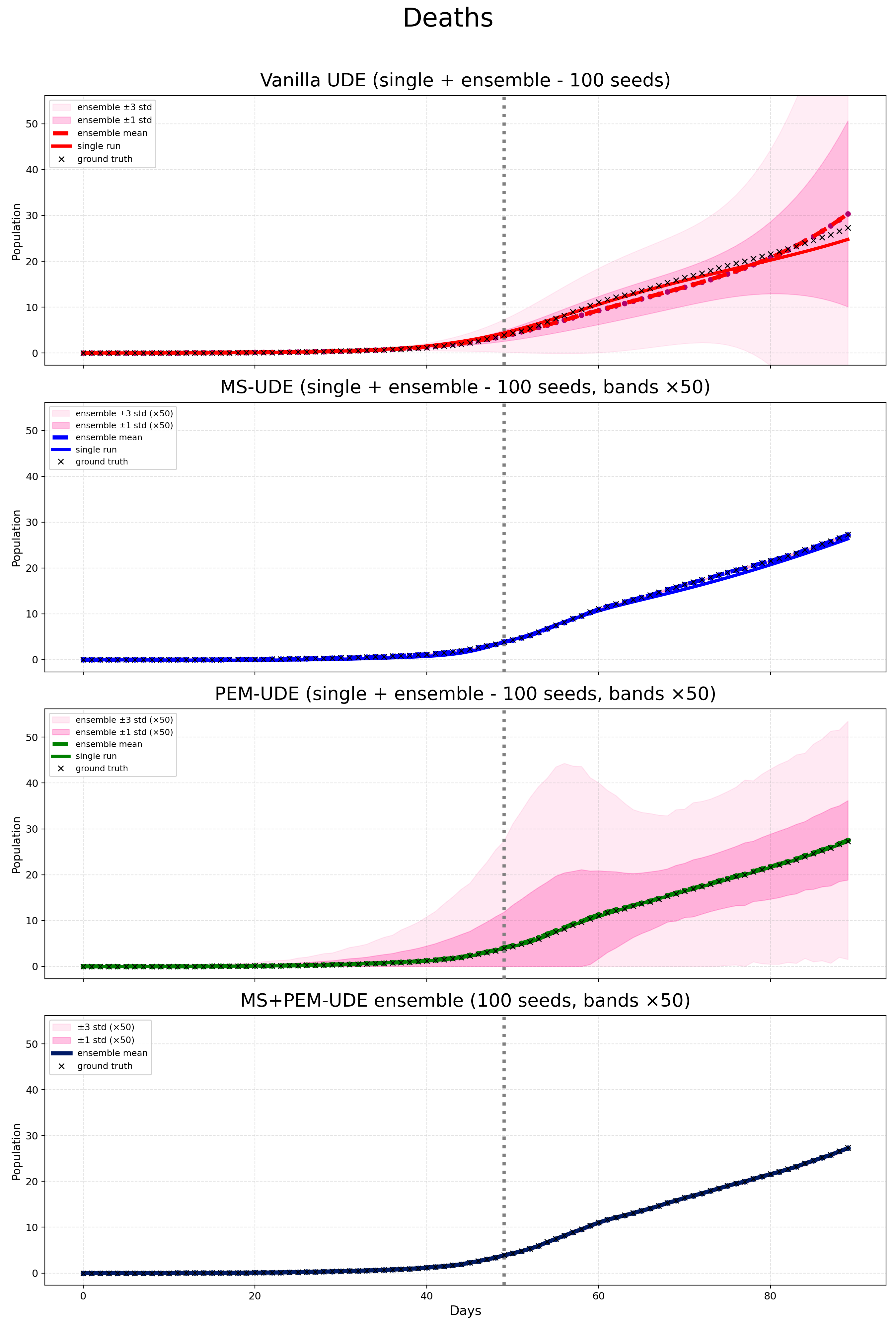}
    \caption{Deaths ($D$).}\label{fig:grid4x1_deaths}
  \end{subfigure}\hfill
  \begin{subfigure}[t]{0.31\textwidth}
    \includegraphics[width=\linewidth]{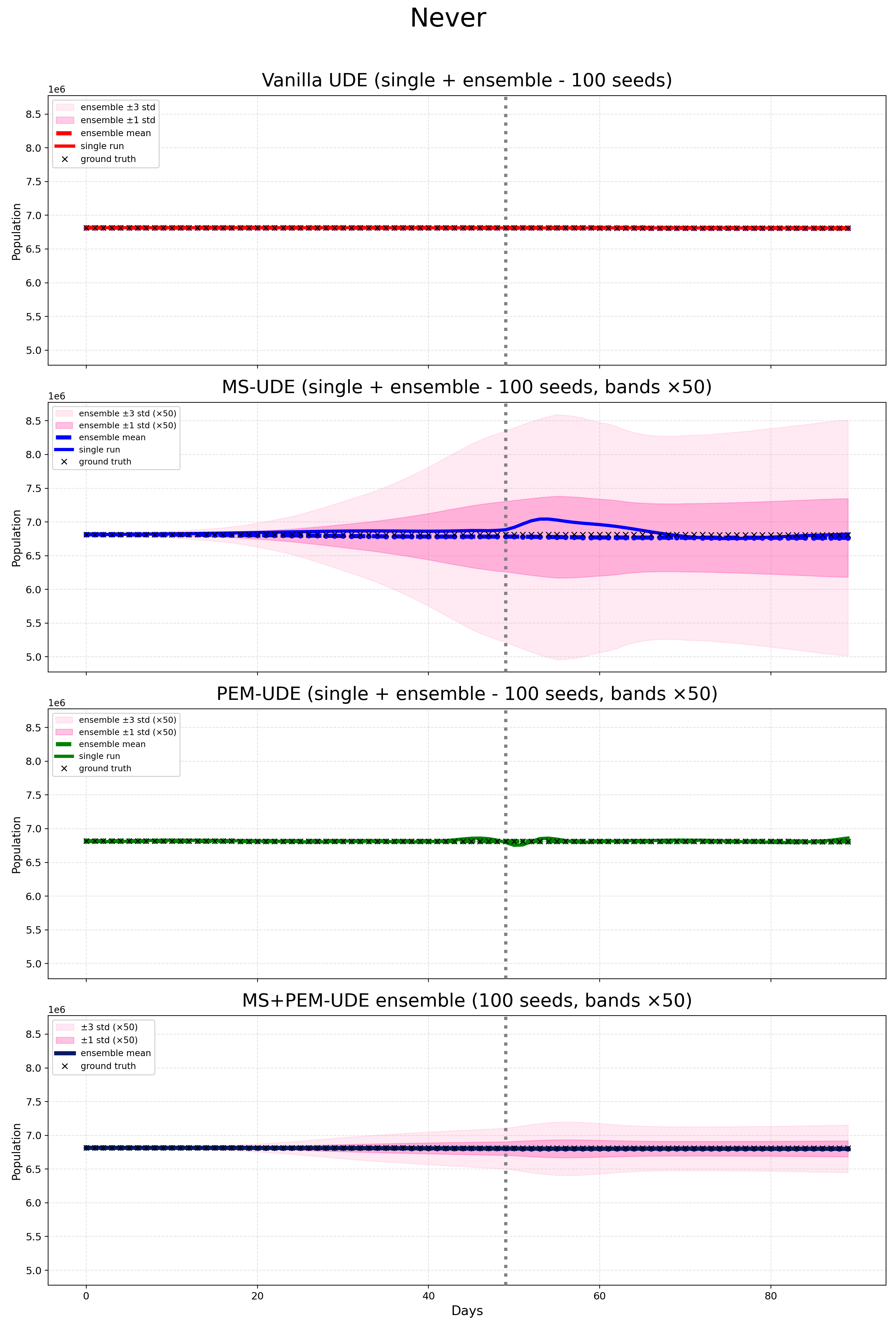}
    \caption{Never-infected / susceptible-like pool.}\label{fig:grid4x1_never}
  \end{subfigure}

  \captionof{figure}{\textbf{Trajectory fit and ensemble robustness (Dataset 1, 100 seeds).}
  Each panel stacks four rows: \emph{(top)} Vanilla UDE, \emph{(second)} MS-UDE, \emph{(third)} PEM-UDE,
  and \emph{(bottom)} MS+PEM-UDE. Black \texttt{x} markers denote ABM ground truth. Shaded regions indicate
  $\pm1\sigma$ and $\pm3\sigma$ variability across 100 independently trained surrogates (random initialization
  and optimizer stochasticity); band scaling (e.g., ``bands $\times 50$'') is annotated in the plots.
  The vertical dotted line marks intervention onset at day 49. MS+PEM-UDE is reported as ensemble mean and
  uncertainty bands (no single-seed trace).}
  \label{fig:grid4x1_ms_pem}
\end{center}

\subsection{Uncertainty Alignment and Calibration}
\label{sec:uncertainty}

Reliable decision support requires calibrated uncertainty, not only accurate means. We evaluate calibration by
computing empirical coverage of surrogate trajectories within ABM quantile envelopes (10--90\% and 25--75\%)
computed across stochastic ABM realizations of Dataset~1. Table~\ref{tab:coverage_summary} reports mean empirical
coverage. MS+PEM-UDE achieves the highest coverage (0.94 / 0.69), while PEM-UDE is the strongest among the single
stabilization variants (0.86 / 0.61), indicating improved distributional alignment without overly diffuse forecasts.

\begin{table}[t]
  \centering
  \caption{Mean empirical coverage of the 10--90\,\% and 25--75\,\% ABM quantile bands on Dataset 1. Higher values
  indicate better calibration to the ensemble variability.}
  \label{tab:coverage_summary}
  \begin{tabular}{lcc}
    \toprule
    \textbf{Method} & \textbf{10--90\,\% Coverage} & \textbf{25--75\,\% Coverage} \\
    \midrule
    UDE (baseline) & 0.68 & 0.43 \\
    MS-UDE & 0.79 & 0.55 \\
    PEM-UDE & 0.86 & 0.61 \\
    MS+PEM-UDE (ensemble) & \textbf{0.94} & \textbf{0.69} \\
    \bottomrule
  \end{tabular}
\end{table}

\paragraph{Ensemble protocol.}
For each training strategy, we quantify surrogate sensitivity by training an \emph{ensemble} of 100 models with
different random seeds (network initialization and optimizer stochasticity). We report ensemble mean trajectories
and $\pm1\sigma/\pm3\sigma$ variability across seeds; MS+PEM-UDE is reported in ensemble form only.

\subsection{Best method}

\begin{table}[t]
\centering
\caption{Relative change in per-compartment MSE of MS+PEM-UDE compared to PEM-UDE, MS-UDE, and Vanilla UDE on Dataset 1. Negative values indicate MS+PEM-UDE achieves lower MSE.}
\label{tab:mspem_all_deltas}
\small
\setlength{\tabcolsep}{4pt}
\renewcommand{\arraystretch}{1.15}
\begin{tabularx}{\linewidth}{>{\RaggedRight\arraybackslash}X c c c}
\toprule
\textbf{Compartment} &
\textbf{MS+PEM vs.\ PEM} &
\textbf{MS+PEM vs.\ MS} &
\textbf{MS+PEM vs.\ UDE} \\
\midrule
Exposed ($E$) & $-2.3\%$ & $-8.6\%$ & $-86.3\%$ \\
Presymptomatic ($I_{\sim s}$) & $-3.4\%$ & $-7.6\%$ & $-82.6\%$ \\
Symptomatic ($I_s$) & $-5.6\%$ & $-10.4\%$ & $-91.2\%$ \\
Asymptomatic ($I_a$) & $-6.8\%$ & $-9.1\%$ & $-87.4\%$ \\
Never-infected / Susceptible-like pool ($S$) & $-1.5\%$ & $-5.4\%$ & $-70.1\%$ \\
Deaths ($D$) & $-4.3\%$ & $-7.3\%$ & $-68.4\%$ \\
\bottomrule
\end{tabularx}
\end{table}

Across Dataset~1, \textbf{MS+PEM-UDE} is the strongest overall method when considering (i) trajectory error,
(ii) robustness to random initialization, and (iii) uncertainty calibration. Figure~\ref{fig:grid4x1_ms_pem}
shows that MS+PEM-UDE produces the most concentrated 100-seed ensemble trajectories while remaining centered
on the ABM ground truth through the intervention and late-horizon dynamics. Quantitatively, Table~\ref{tab:mspem_all_deltas}
confirms that MS+PEM-UDE improves per-compartment MSE relative to \emph{all} baselines: it consistently reduces
error versus PEM-UDE and MS-UDE across every compartment, and achieves large relative reductions versus the
single-shooting UDE baseline. These gains are largest in the infectious compartments ($E$, $I_{\sim s}$, $I_s$,
$I_a$), where long-horizon drift and phase error are most pronounced for single shooting. MS+PEM-UDE also ends up with the lowest loss (\ref{fig:loss-plot})

In addition to lower error, MS+PEM-UDE yields the best reliability among the evaluated surrogates, achieving
the highest empirical coverage within ABM quantile bands (Table~\ref{tab:coverage_summary}). Together, these
results support MS+PEM-UDE as the best-performing surrogate for Dataset~1, combining mechanistic structure with
stabilized training that improves both accuracy and risk-aware calibration.

\section{Discussion}
\label{sec:discussion}

This study evaluates stabilized UDE surrogates for a single representative ABM scenario (Dataset~1) using
\textbf{100-seed training ensembles} to quantify robustness to random initialization and optimization
stochasticity. Within this setting, stabilization consistently improves both trajectory fidelity and
distributional reliability relative to the single-shooting baseline. MS-UDE reduces long-horizon drift by
shortening the effective backpropagation horizon through windowed training, PEM-UDE further stabilizes
identification by re-anchoring states via an observer correction, and the combined method, \textbf{MS+PEM-UDE},
yields the best overall performance. In particular, MS+PEM-UDE achieves the lowest overall trajectory error
on Dataset~1 (MSE $=2.93$ vs.\ $3.01$ for PEM-UDE, $3.76$ for MS-UDE, and $16.85$ for the baseline) while
maintaining tight ensemble dispersion and strong alignment to ABM ground truth in the ensemble visualization
(Fig.~\ref{fig:grid4x1_ms_pem}). These findings support \textbf{Hypothesis~H2} in the reported setting: stabilized
UDE surrogates can reconstruct ABM aggregate dynamics over decision-relevant horizons while preserving the
mechanistic structure of the compartment model.

Improved calibration has direct operational value. For hospital planners, uncertainty estimates are as important
as point forecasts: under-dispersion leads to overconfidence and missed surge risk, while excessive dispersion
reduces actionable signal. Using ABM quantile envelopes as a reference, empirical coverage improves monotonically
with stabilization on Dataset~1, rising from 0.68/0.43 (UDE) to 0.79/0.55 (MS-UDE), 0.86/0.61 (PEM-UDE), and
\textbf{0.94/0.69} (MS+PEM-UDE) for the 10--90\% and 25--75\% bands (Table~\ref{tab:coverage_summary}). This
indicates that the most stable training strategy also yields the most reliable uncertainty alignment to the
stochastic variability of the underlying ABM.

From a deployment perspective, the surrogate achieves practical runtimes on commodity hardware. Inference for a
$\sim$90-day forecast runs in tens of seconds on a laptop CPU (Table~\ref{tab:inference_m4}), enabling rapid
``what-if'' evaluation without re-running expensive ABM simulations. A realistic workflow would integrate the
surrogate into a lightweight tool where staff input recent observed counts (or a mapped subset under a realistic
observation operator $H$), select a short horizon (e.g., 1--3 weeks), and explore intervention timing/strength
to assess threshold-driven outcomes (e.g., ICU occupancy approaching 75\%). Each run can log model version,
timestamp, and random seed for provenance, supporting auditability in high-stakes settings.

\paragraph{Limitations and future work.}
This report is intentionally centered on one scenario (Dataset~1) because robust ensemble evaluation (100 seeds)
is computationally costly. Extending the same pipeline to additional ABM scenarios is an important next step to
establish generality across intervention regimes. In addition, the current study uses synthetic ABM-generated
compartment trajectories; real surveillance data introduce partial observability, reporting delays, and
measurement noise. Future work should therefore evaluate training under a realistic $H$ (observing only subsets
such as admissions, ICU, or deaths), include noise models, and compare against external data when available.
Finally, while the observer gain $K$ is constrained for stability (e.g., diagonal / block-diagonal), richer
structures could improve responsiveness but require careful regularization and ablations under equalized
training budgets.

\section{Conclusion}
\label{sec:conclusion}

We developed a mechanistically constrained ABM$\rightarrow$UDE surrogate for epidemic dynamics and evaluated
stabilized training strategies on a single representative ABM scenario (Dataset~1) using \textbf{100-seed
training ensembles}. Compared to the single-shooting baseline, stabilization substantially improves both
trajectory fidelity and reliability. MS+PEM-UDE is the best-performing method in this study: it achieves the
lowest overall trajectory error on Dataset~1 (MSE $=2.93$ vs.\ $3.01$ for PEM-UDE, $3.76$ for MS-UDE, and
$16.85$ for UDE) and the highest uncertainty calibration against ABM variability (coverage
\textbf{0.94/0.69} for the 10--90\% / 25--75\% bands). Inference runs in tens of seconds on commodity hardware
(approximately 21--32\,s per $\sim$90-day forecast), enabling rapid counterfactual evaluation without requiring
HPC-scale resources.

These results demonstrate that a stabilized UDE surrogate can compress ABM realism into an operationally usable
model while preserving mechanistic interpretability and providing risk-aware uncertainty. Future work will
extend ensemble evaluation to additional ABM scenarios, incorporate partial-observability training under
realistic measurement operators, and validate against real-world hospital or public health time series when
available.

\bibliography{rebs}

\end{document}